\documentclass[runningheads]{llncs}

\usepackage{eccv}

\usepackage{eccvabbrv}

\usepackage{graphicx}
\usepackage{booktabs}
\usepackage{multirow}

\usepackage[accsupp]{axessibility}  

\usepackage{hyperref}

\usepackage{orcidlink}

\begin{document}

\title{ICDepth: Taming Video Diffusion Models for Video Depth Estimation via In-Context Conditioning} 

\titlerunning{ICDepth}

\author{
Xuanhua He\textsuperscript{*} \and
Jiaxin Xie\textsuperscript{*} \and
Mingzhe Zheng \and
Qifeng Chen\textsuperscript{\dag}
}
\begingroup
\renewcommand{\thefootnote}{\fnsymbol{footnote}}
\footnotetext[1]{Equal contribution. \quad \textsuperscript{\dag} Corresponding author.}
\endgroup
\authorrunning{X. He et al.}

\institute{
The Hong Kong University of Science and Technology, Hong Kong SAR, China
}
\maketitle

\begin{center}
\vspace{-1.2em}
\small
\textbf{Project Page:} \href{https://xuanhuahe.github.io/ICDepth/}{https://xuanhuahe.github.io/ICDepth/}
\end{center}
\vspace{-0.8em}

\begin{abstract}
Monocular video depth estimation requires temporal consistency, geometric accuracy, and generalization across diverse scenarios—yet existing methods 
struggle to achieve all three simultaneously. Discriminative models excel 
at per-frame accuracy but suffer from temporal drift due to limited context 
windows, while generative methods improve consistency and generalization at the cost of extensive training data (10M+ samples) and lack of geometric precision. In response to these issues, we introduce \textbf{ICDepth}, a framework that adapts pre-trained text-to-video diffusion transformers for video depth estimation via In-Context Conditioning (ICC), leveraging their rich spatial-temporal priors. To address key challenges in transferring ICC from generation to dense prediction, we propose: (1)~\textbf{SAND-Attention}, which ensures precise spatial-temporal alignment via shared RoPE and enforces unidirectional attention to prevent noise contamination; (2)~\textbf{SRFM}, which injects DINOv2 semantic and resolution priors to enhance geometric precision. 
ICDepth achieves state-of-the-art results on multiple benchmarks with remarkable data efficiency, trained on only 0.8M frames (6–13$\times$ less than competing generative methods), while demonstrating strong zero-shot generalization to diverse domains.
\end{abstract}

\section{Introduction}
\label{sec:intro}
\begin{figure*}
    \centering
    \includegraphics[width=\linewidth]{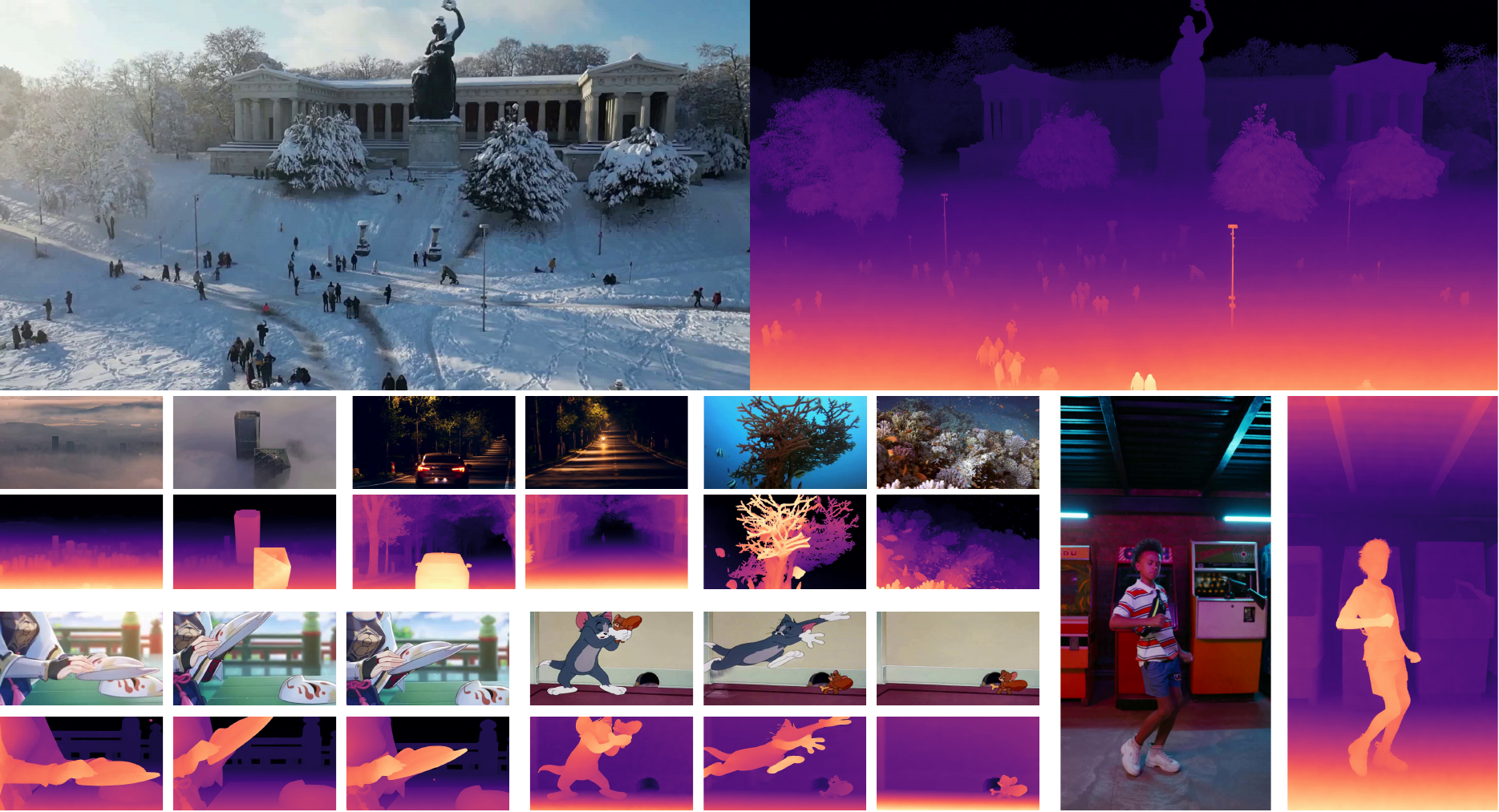}
    \caption{Robust depth estimation across diverse scenarios. Our method performs high-resolution depth estimation (1080p) on videos with varying aspect ratios and demonstrates strong generalization across challenging conditions including foggy weather, nighttime scenes, underwater footage, and both 2D and 3D animated content.}
    \label{fig:first_fig}
\end{figure*}

Monocular video depth estimation, which predicts a dense depth map for each frame in a video, is a fundamental task in 3D computer vision. It is crucial for many applications, such as AR/VR, autonomous driving, and 3D reconstruction. While recent models like Depth Anything V2~\cite{yang2024depth} have made great progress in single-image depth estimation, video-based methods face three core challenges. The first is maintaining temporal consistency to ensure the depth maps are stable and flicker-free across frames. The second is effectively leveraging temporal information from adjacent frames to improve the accuracy and robustness of the prediction for the current frame. The third is generalization ability in diverse scenes. Overcoming these challenges, especially in long videos with complex motion, remains a primary goal in this field.

Current approaches to video depth estimation predominantly fall into two categories. Discriminative models, such as Video Depth Anything (VDA)~\cite{chen2025video}, typically adapt powerful single-image backbones by appending a post-processing temporal head. However, this strategy is inherently limited; the temporal head usually operates over a narrow local window, struggling to capture long-range dependencies. This constrained temporal modeling capacity causes the consistency drift in long videos. The results of discriminative models are also overly smooth, lacking fine-grained details and their limited pre-training on diverse video datasets restricts generalization capability.
On the other hand, generative models, like DepthCrafter~\cite{hu2025depthcrafter} and Depth Any Video~\cite{yang2024deptha}, leverage pre-trained video diffusion models, primarily Stable Video Diffusion (SVD)~\cite{blattmann2023stable}. However, their performance is constrained by two factors. First, the foundational SVD model itself is a U-Net architecture with relatively weak prior~\cite{peebles2023scalable}. Second, they employ a factorized spatio-temporal approach, interleaving 3D convolutional layers with separate 1D temporal attention modules. Adapting them for depth estimation often requires invasive architectural modifications and extensive training on massive datasets (more than 10M samples).

Discriminative methods excel at geometric accuracy but fail at generalization and cannot effectively utilize long-range information, while generative methods improve temporal consistency but sacrifice accuracy and demand prohibitive data scale. Can we train a model that satisfies all three requirements with limited data?

Recent text-to-video diffusion transformer models, such as Wan~2.1~\cite{wan2025wan}, offer a promising alternative. Through training on millions of diverse videos, these models have learned rich scene and spatial-temporal priors for generating videos. Unlike factorized models, VDiT employ native 3D attention that jointly processes all spatial and temporal tokens, naturally capturing complex spatio-temporal dependencies. This raises a question: \textit{Can we transfer these temporal priors to depth estimation while adding the geometric precision that generative models lack?}

However, standard conditioning approaches are ill-suited for VDiT. Concatenating RGB and depth features along the channel dimension requires modifying the input projection layer and provides only weak feature interaction through separate channels. Instead, we adopt \textit{In-Context Conditioning} (ICC)~\cite{ju2025fulldit,he2025fulldit2}, which treats RGB and depth as a unified token sequence $\mathbf{s} = [\mathbf{z}_{\mathrm{depth}}; \mathbf{c}_{\mathrm{RGB}}]$ processed by the VDiT's native attention. This design is non-invasive and provides rich cross-modal interactions through attention.

Transferring the ICC paradigm from generation to a dense prediction task like depth estimation introduces two non-trivial challenges, arising from ICC itself and the feature space of VDiT. First, in the vanilla ICC framework, the noisy depth latent $\mathbf{z}_{\text{depth}}$ would corrupt the clean RGB condition through bidirectional attention, degrading conditioning quality in deeper layers. Second, T2V models are trained for generation and lack both geometric precision and explicit resolution awareness needed for accurate multi-resolution depth prediction.


To address these challenges, we propose ICDepth, a specialized ICC framework for video depth estimation built upon two key innovations. First, we introduce SAND-Attention (Spatial-temporal Aligned, Noise-Decoupled Attention) to resolve noise contamination. This mechanism establishes precise spatial-temporal alignment between clean RGB conditions and noisy depth latents through RoPE Alignment, while ensuring strict unidirectional information flow by Decoupled Attention. The Decoupled Attention operates in a specialized two-stage process: noisy depth tokens can query clean RGB tokens for contextual guidance, while reverse attention is systematically blocked. Additionally, we zero-out timestep embeddings for RGB conditions, effectively shielding the clean conditioning signal from temporal noise interference.

Furthermore, to tackle geometric ambiguity, we introduce a Semantic- and Resolution-Aware Feature Modulation Block (SRFM). This module injects powerful external priors into the generative model, leveraging DINOv2~\cite{oquab2023dinov2} features for rich semantic understanding and resolution embeddings for multi-scale inference. These innovations transform the generative model into a precise depth estimation foundation. Remarkably, our model achieves SOTA performance across diverse datasets despite training on a small-scale synthetic dataset (only 0.8M frames). As shown in Figure~\ref{fig:first_fig}, our method produces highly accurate depth predictions while demonstrating strong generalization across diverse scenarios—including foggy weather, nighttime scenes, underwater footage, and animated content—and flexibly handles varying resolutions and aspect ratios up to 1080p.

Our contributions are summarized as follows:
\begin{itemize}
    \item We introduce the In-Context Conditioning (ICC) paradigm for video depth estimation, enabling a powerful text-to-video diffusion transformer to serve as a state-of-the-art depth estimation model.
\item We propose a comprehensive solution to overcome the key challenges of applying ICC to this new domain. Our solution includes: a novel SAND-Attention mechanism that resolves the noise contamination issue via shared RoPE and one-way attention; and a semantic- and resolution-aware feature modulation block that injects DINOv2 and resolution priors to tackle the model's inherent geometric ambiguity.
\item Our framework achieves state-of-the-art performance on multiple benchmarks with remarkable data efficiency.
\end{itemize}
\section{Related Work}
\label{sec:related_work}
\subsection{Video Depth Estimation}
Video depth estimation aims to generate temporally consistent depth sequences~\cite{luo2020consistent,zhang2019exploiting,kopf2021robust}. Early approaches often relied on Test-Time Optimization, which fine-tunes a pre-trained single-image model for each new video~\cite{kopf2021robust,luo2020consistent,zhang2021consistent}. While capable of producing consistent results, these methods have high inference overhead and are often dependent on precise camera poses or optical flow, limiting their applicability in the wild. Feed-forward methods predict depth directly from video clips. Some approaches introduce plug-and-play modules to stabilize off-the-shelf monocular depth predictions, such as NVDS~\cite{wang2023neural}. Others leverage geometric priors like optical flow or poses to enforce consistency. However, the performance of these models is often susceptible to errors in the priors, and their generalization is frequently hampered by the scarcity of diverse, large-scale video depth training data.
More recently, a new line of work leverages the powerful priors of pretrained model such as Depth Anything V2~\cite{yang2024depth} or SVD~\cite{blattmann2023stable}. Concurrent works such as ChronoDepth~\cite{shao2025learning}, DepthCrafter~\cite{hu2025depthcrafter}, and Depth Any Video~\cite{yang2024deptha} repurpose these generative models to produce high-fidelity and temporally coherent depth sequences, while Video Depth Anything~\cite{chen2025video} designs a temporal head~\cite{kuang2025buffer} to leverage the prior from Depth Anything V2.
\subsection{In-Context Conditioning for Diffusion Models}
The field has witnessed a paradigm shift toward leveraging sequence concatenation as a mechanism for In-Context Conditioning within diffusion transformers~\cite{huang2024context,tan2024ominicontrol,wu2025less,li2025visualcloze,song2025insert,mou2025dreamo,ma2025controllable,ma2025followcreation}. Pioneering work such as Omnicontrol~\cite{tan2024ominicontrol} established the foundational concept of unifying conditioning signals and noisy latent representations within a single sequential structure. Through the inherent self-attention operations of Transformers, these combined sequences enable implicit learning of conditioning dependencies. The architectural simplicity of this methodology presents a compelling advantage—achieving sophisticated multi-modal control while preserving the original DiT framework, consequently enhancing both controllability and synthesis fidelity.
This concatenation-based conditioning strategy has catalyzed a wave of follow-up investigations in image synthesis~\cite{wu2025less,li2025visualcloze,song2025insert,mou2025dreamo}. More recently, the paradigm has been successfully adapted to address the inherently more complex video generation task, demonstrating remarkable efficacy~\cite{ju2025fulldit,lin2025omnihuman,guo2025long,ye2025unic,he2025fulldit2,ma2024followpose,ma2025followcreation,ma2026fastvmt}.

\section{Method}
\label{sec:method}
In this section, we first briefly introduce the concept of video diffusion model, then we dive into the details of our proposed method. 
\begin{figure*}
    \centering
    \includegraphics[width=\linewidth]{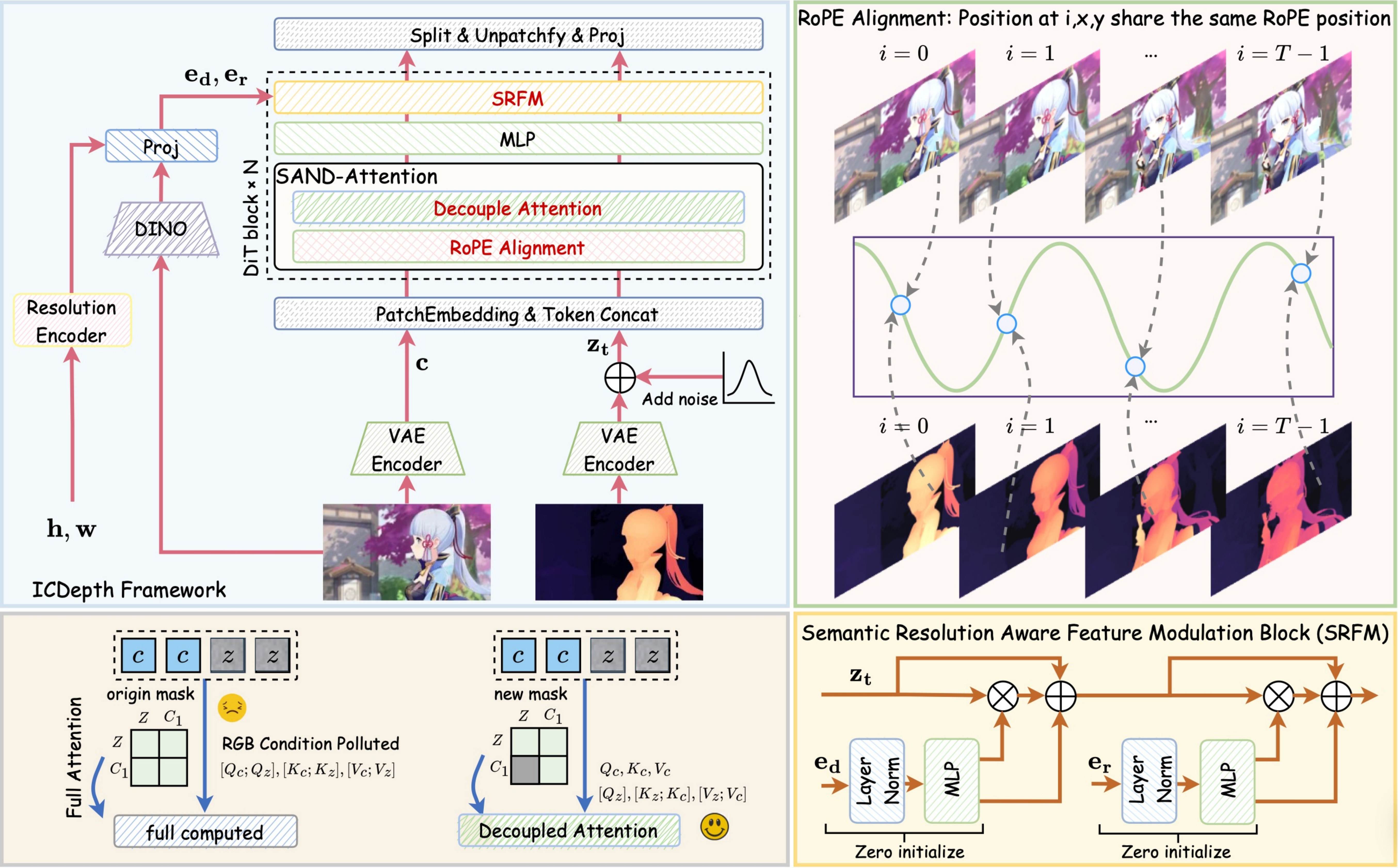}
    \caption{The framework of our proposed ICDepth, which contains two components: SRFM and SAND-Attention. The RoPE alignment ensures that corresponding spatial-temporal positions in the video-depth paired data share identical RoPE positional encodings.}
    \label{fig:mainfig}
\end{figure*}
\subsection{Preliminary}
\noindent\textbf{video diffusion transformer (VDiT) } Our approach builds upon the VDiT, which leverages a pure transformer architecture for video generation. The model $u_\Theta$ is composed of stacked transformer blocks that jointly implement \textbf{spatial-temporal self-attention} and \textbf{cross-attention}, followed by \textbf{MLPs}.

This design provides a unified architecture for capturing complex, long-range spatio-temporal dependencies, as opposed to hybrid CNN-attention models like UNet. The self-attention mechanism globally relates all patches in the spatio-temporal volume, while the cross-attention seamlessly integrates conditional information at every layer.

Following the Flow Matching framework~\cite{lipman2022flow}, we train the model to predict the velocity field $\mathbf{v}_t$ through the objective:
\begin{equation}
\label{eq:flow_matching}
\mathcal{L} = \mathbb{E}_{t, \mathbf{z}_0, \mathbf{z}_1, \mathbf{c}} \| u_\Theta(\mathbf{z}_t, t, \mathbf{c}) - \mathbf{v}_t \|^2
\end{equation}
where $u_\Theta$ represents our VDiT backbone, $\mathbf{z}_t$ denotes the noisy latent video representation at time $t$, and $\mathbf{c}$ encompasses conditional inputs.

During inference, we generate video samples by solving the corresponding ODE defined by the learned velocity field using numerical solvers:
\begin{equation}
\frac{d\mathbf{z}(t)}{dt} = u_\Theta(\mathbf{z}_t, t, \mathbf{c})
\end{equation}
This formulation enables efficient sampling while maintaining temporal coherence across video frames.

\subsection{Overview}
\label{sec:overview}
The framework of our proposed ICDepth is shown in Figure~\ref{fig:mainfig}. ICDepth builds upon the pretrained text -to-video generation model Wan~\cite{wan2025wan} with two core designs for video depth estimation.

Given an input RGB video $V_I$ and its corresponding depth video $V_D$, we first encode each modality into the latent space using a pretrained VAE encoder. This yields the clean latent representation $\mathbf{z}_0$ from $V_D$ and the conditioning latent $\mathbf{c}$ from $V_I$. We then corrupt $\mathbf{z}_0$ by adding Gaussian noise $\epsilon \sim \mathcal{N}(\mathbf{0}, \mathbf{I})$ to obtain the noisy latent $\mathbf{z}_t$ at diffusion time step $t$. Both $\mathbf{z}_t$ and $\mathbf{c}$ are patchified and reshaped into tensors of shape $\mathbb{R}^{n \times c}$, then concatenated along the token dimension to form the unified sequence $\mathbf{s}_t = [\mathbf{z}_t; \mathbf{c}] \in \mathbb{R}^{2n \times c}$.

To incorporate prior knowledge into the denoising model, we introduce both semantic and resolution information to enrich the generation process. For semantic conditioning, we extract DINOv2 features~\cite{oquab2023dinov2} from the input video $V_I$, obtaining feature maps $f_d$ that are subsequently pooled and reshaped into semantic embeddings $\mathbf{e}_d$. For resolution conditioning, the spatial dimensions $(h, w)$ are encoded to $(\mathbf{e}_h, \mathbf{e}_w)$ using sinusoidal positional embeddings, which are concatenated and processed through MLP to yield the resolution embedding $\mathbf{e}_r$.

These conditioning signals are fed into the VDiT model, which is trained to estimate the vector field that characterizes the transformation between the noise distribution and the data distribution. The overall objective is to minimize the flow matching loss. During inference, depth estimation is achieved by solving the ODE defined by the learned vector field, starting from Gaussian noise and conditioned on the input RGB video with its derived semantic and resolution features, generating accurate and temporally consistent depth maps through numerical integration.

\subsection{Key Component}
\subsubsection{In-Context Conditioning}
To adapt a pretrained video generation model for depth estimation, we need to effectively incorporate the RGB video condition. Existing methods~\cite{ke2024repurposing,hu2025depthcrafter,yang2024deptha} achieve this by concatenating $\mathbf{z}$ and $\mathbf{c}$ along the channel dimension, followed by fusion through projection layers. However, this approach has limitations: it causes information loss and underutilizes the attention mechanisms of the pretrained model, as channel concatenation and projection cannot establish strong relationships between the two types of tokens.
Instead, we leverage the ICC framework, where the condition $\mathbf{c}$ is provided as context tokens without information loss. The model learns the relationships between depth and RGB features within the attention layers. Specifically, by concatenating $\mathbf{z}$ and $\mathbf{c}$ along the token dimension to form $\mathbf{s}=[\mathbf{z};\mathbf{c}]$, the self-attention mechanism naturally models their interactions. This framework enables the model to fully exploit the pretrained attention capabilities, leading to improved performance.

\subsubsection{SAND-Attention}
Traditional ICC framework processes $\mathbf{s}=[\mathbf{z};\mathbf{c}]$ through full self-attention, suffering from three key issues: (1) RoPE creates artificial sequential relationships between aligned depth-RGB tokens; (2) Bidirectional attention allows noise from $\mathbf{z}$ to degrade $\mathbf{c}$; (3) Uniform timestep embedding $\mathbf{e}_t$ injects unnecessary variance into stable RGB features.

To address these limitations, we propose \textbf{SAND-Attention} (Spatial-temporal Aligned and Noise-Decoupled Attention). We first separate queries, keys, and values for depth and conditioning tokens:
\begin{equation}
    Q_{\mathbf{z}},Q_{\mathbf{c}} = \text{split}(Q_{\mathbf{s}}), \quad
    K_{\mathbf{z}},K_{\mathbf{c}} = \text{split}(K_{\mathbf{s}}), \quad
    V_{\mathbf{z}},V_{\mathbf{c}} = \text{split}(V_{\mathbf{s}})
\end{equation}

We then apply RoPE with shared positional indices $\mathcal{P}$ to ensure spatial-temporal alignment:
\begin{align}
    Q'_\mathbf{z},Q'_\mathbf{c} &= \text{RoPE}(Q_\mathbf{z},\mathcal{P}), \text{RoPE}(Q_\mathbf{c},\mathcal{P}) \nonumber\\
    K'_\mathbf{z},K'_\mathbf{c} &= \text{RoPE}(K_\mathbf{z},\mathcal{P}), \text{RoPE}(K_\mathbf{c},\mathcal{P})
\end{align}

The core innovation is our noise-decoupled attention mechanism. We compute self-attention within clean conditioning tokens:
\begin{equation}
    O_\mathbf{c} = \mathrm{softmax}\left(Q'_\mathbf{c}{K'_\mathbf{c}}^{T}\right)V_\mathbf{c}
\end{equation}
Followed by unidirectional cross-attention from noisy depth to clean conditioning:
\begin{equation}
    O_\mathbf{z}  = \text{softmax}(Q'_\mathbf{z}[K'_\mathbf{z};K'_\mathbf{c}]^T)[V_\mathbf{z};V_\mathbf{c}]
\end{equation}

Finally, we concatenate outputs $O_\mathbf{s}=[O_\mathbf{z};O_\mathbf{c}]$ and modify timestep conditioning by zeroing $\mathbf{e}_t$ for $\mathbf{c}$ tokens, preserving RGB feature stability across diffusion steps. This design maintains compatibility with flash attention while ensuring clean conditioning information flows unidirectionally to guide depth generation.

\subsubsection{Semantic-Resolution Aware Feature Modulation}
Although the pretrained VDiT model can benefit from the ICC framework and our SAND-Attention, certain limitations remain. While the T2V model is trained on extremely large-scale datasets, its learned representations are not always optimal for perception tasks such as depth estimation. Moreover, depth maps are highly correlated with image resolution and aspect ratio. During inference, different test videos have varying aspect ratios, and using a fixed resolution leads to performance degradation. 

To address these issues, we enhance the VDiT model using Semantic-Resolution Aware Feature Modulation Block (SRFM). First, we incorporate DINOv2~\cite{oquab2023dinov2} features as prior prompts to enrich the semantic feature representation. Second, we provide explicit resolution information to improve performance across different training and inference resolutions, particularly for non-standard aspect ratios.

Given feature $s^{l}_{\text{mlp}}$ after the MLP layer in a VDiT block, we inject DINOv2 and resolution priors. As described in Section~\ref{sec:overview}, we have embeddings $e_d$ and $e_r$ representing DINOv2 and resolution priors, respectively. We project them using two MLP-based functions:
\begin{align}
    e^{l}_d,e^{l}_r = \Theta_d(e_d),\Theta_r(e_r)
\end{align}
where the feature channels are doubled through this projection. We then split the features along the channel dimension to obtain scale and shift parameters for modulation:
\begin{equation}
    e^{\text{scale}}_d, e^{\text{shift}}_d = \text{split}(e^{l}_d) \qquad e^{\text{scale}}_r, e^{\text{shift}}_r = \text{split}(e^{l}_r)
\end{equation}
To inject this information into $\mathbf{z}$, we split $\mathbf{s}^l$ into $\mathbf{z}^l$ and $\mathbf{c}^l$, then apply sequential modulation:
\begin{align}
    \mathbf{z}^{l}_d = \mathbf{z}^l \cdot (1+e^{\text{scale}}_d)+e^{\text{shift}}_d, & \quad \mathbf{z}^{l}_r = \mathbf{z}^{l}_d \cdot (1+e^{\text{scale}}_r)+e^{\text{shift}}_r \nonumber \\
    s^{l+1} &= [\mathbf{z}^{l}_r;\mathbf{c}^l]
\end{align}
This design improves model performance across diverse scenes and resolutions.
\label{sec:experiment}
\subsection{Training Pipeline}

Given an input RGB video $V_I$ and its corresponding depth video $V_D$,
we encode $V_D$ into the clean depth latent $z_0$ using the VAE encoder,
while $V_I$ is encoded as the conditioning latent $c$.
For the corresponding depth video $V_D \in \mathbb{R}^{f\times h \times w}$, we first compute a binary mask $M \in \{0,1\}^{f\times h \times w}$ by thresholding against the maximum depth value $D_{\text{max}}$:
\[
M_{i,j,k} = \begin{cases}
1 & \text{if } V_D(i,j,k) < D_{\text{max}} \\
0 & \text{otherwise}
\end{cases}
\]
where the mask identifies valid depth regions.

For relative depth estimation training, we convert depth $V_D$ to disparity and normalize it to range $[-1,1]$ by per-video min-max normalization. The mask $M$ is adapted to the VAE latent resolution.

We train the model using the flow matching objective, where the loss is computed only in valid regions ($M=1$):
\begin{equation}
    \mathcal{L} = \mathbb{E}_{t, \mathbf{z}_0, \mathbf{z}_1, \mathbf{c},\mathbf{e}_r,\mathbf{e}_d}\left[ {M} \odot \| \mathbf{u}_{\Theta}(\mathbf{z}_t, t, \mathbf{c},\mathbf{e}_r,\mathbf{e}_d) - \mathbf{v}_t \|^2 \right]
\end{equation}
where $\odot$ denotes element-wise multiplication.

\section{Experiment}

\subsection{Dataset and Benchmark}
Compared to DepthCrafter~\cite{hu2025depthcrafter}, we train our model on a more compact dataset comprising: (1) VKITTI~\cite{cabon2020virtual}, (2) subsets of TartanAir~\cite{wang2020tartanair} and TartanGround~\cite{patel2025tartanground} (using single-direction cameras), and (3) the synthetic subset of OmniWorld~\cite{zhou2025omniworld}. The total training datasets contain around 0.8M frames. 
For relative depth estimation, we compare against state-of-the-art video and image depth estimation methods, including both discriminative and generative models: ChronoDepth~\cite{shao2025learning}, DepthCrafter~\cite{hu2025depthcrafter}, Depth Any Video~\cite{yang2024deptha}, Video Depth Anything~\cite{chen2025video}, and Depth Anything V2~\cite{yang2024depth}.
We evaluate on diverse benchmarks encompassing both synthetic and real-world scenarios, with video lengths ranging from 50 to over 100 frames. Specifically, we use Sintel~\cite{mayer2016large}, KITTI~\cite{geiger2013vision}, ScanNet~\cite{dai2017scannet} and Bonn~\cite{bonn} for quantitative evaluation. Additionally, we provide qualitative comparisons on challenging domains including animation, gaming, and underwater scenes to demonstrate our method's generalization capability.
We evaluate our results using three standard metrics: Root Mean Square Error (RMSE), threshold accuracy ($\delta1$), and Absolute Relative error (AbsRel).

It is worth noting that the reported performance of these baseline methods often varies across different publications due to discrepancies in their adopted evaluation protocols. To ensure a fair comparison, we conduct inference for all baseline methods and evaluate their predictions using the official evaluation script provided by DepthCrafter~\cite{hu2025depthcrafter}. Consequently, our evaluated metrics for these baselines may differ from those reported in their original papers, appearing either lower or higher depending on the specific method and dataset.
\subsection{Implementation Details}
We build our model upon the pretrained Wan2.1 1.3B T2V model~\cite{wan2025wan}. Training is conducted on 4 H800 GPUs for 8 epochs with a batch size of 1 per GPU, a learning rate of $2\times10^{-4}$, and gradient accumulation over 32 steps.
We employ multi-resolution training to preserve the original aspect ratios of videos while ensuring spatial dimensions are divisible by 32. The temporal length varies from 21 to 77 frames depending on the spatial resolution of each video. To maintain a reasonable computational budget, we use $H \times W \times T = 672 \times 384 \times 77$ as our token budget baseline, adjusting temporal length inversely with spatial resolution. Detailed configurations of this strategy are provided in the supplementary material. During inference, we set diffusion sampling steps to 5.

\subsection{Relative Depth Comparison with SOTA Methods}
\subsubsection{Quantitative Comparison}
We evaluate different methods through zero-shot video depth estimation experiments, with quantitative results presented in Table~\ref{tab:maintable}. Our method achieves state-of-the-art performance on Sintel, KITTI, and Bonn datasets, obtaining the best results in both AbsRel and $\delta_1$ metrics. Specifically, we achieve significant improvements of 16.0\% in AbsRel and 10.1\% in $\delta_1$ on Sintel compared to the second-best method; Sintel is the benchmark featuring highly dynamic scenes and complex motion patterns. On ScanNet, our method achieves competitive second-best performance, closely matching the top results while using substantially less training data. Notably, our approach accomplishes this superior performance with only 0.8M training samples, representing 6-13$\times$ reduction in data requirements compared to other state-of-the-art video depth estimation methods, demonstrating remarkable data efficiency without compromising estimation accuracy.
\begin{table}[h]
\centering
\caption{Quantitative comparison of zero-shot video depth estimation with state-of-the-art methods on the Sintel, ScanNet, KITTI and Bonn dataset. The \textbf{best} and \underline{second-best} results are highlighted in bold and underlined, respectively. $\downarrow$: lower is better; $\uparrow$: higher is better.}\label{tab:maintable}

\renewcommand{\arraystretch}{1.8} 
\setlength{\tabcolsep}{8pt} 

\large 

\resizebox{\textwidth}{!}{
\begin{tabular}{l cc cc cc cc ll}
\toprule
\multirow{2}{*}{Method} & \multicolumn{2}{c}{Sintel (50 frames)} & \multicolumn{2}{c}{ScanNet (90 frames)} & \multicolumn{2}{c}{KITTI (110 frames)} & \multicolumn{2}{c}{Bonn (110 frames)} & \multirow{2}{*}{Venue} & \multirow{2}{*}{Data size} \\
\cmidrule(lr){2-3} \cmidrule(lr){4-5} \cmidrule(lr){6-7}\cmidrule(lr){8-9}
                        & AbsRel$\downarrow$ & $\delta_1$$\uparrow$ & AbsRel$\downarrow$ & $\delta_1$$\uparrow$ & AbsRel$\downarrow$ & $\delta_1$$\uparrow$ & AbsRel$\downarrow$ & $\delta_1$$\uparrow$ & & \\
\midrule
Depth Anything V2         &      0.403 &      0.547 &      0.123 &      0.852 &      0.102 &      0.910 &      0.084 &      0.947   & NeurIPS'24 &      62.62M\\
Video Depth Anything    &      0.383 &      0.629 &      \textbf{0.075} &      \textbf{0.954} &      \underline{0.078} &      \underline{0.950} &      \textbf{0.053} &      \underline{0.975}     & CVPR'25    &      1.35M\\
\midrule
ChronoDepth             &      0.587 &      0.486 &      0.159 &      0.783 &      0.167 &      0.759 &      0.100 &      0.911 &      CVPR'25    &      -\\
DepthCrafter            &      0.313 &      \underline{0.680} &      0.142 &      0.803 &      0.105 &      0.898 &      0.066 &      0.971 &      CVPR'25    &      10.5M\\
Depth Any Video           &      \underline{0.300} &      0.643 &      0.119 &      0.865 &      0.098 &      0.925 &      0.063 &      0.963 &    ICLR'25    & 6M   \\
\midrule
\textbf{Ours}           &      \textbf{0.250} &      \textbf{0.749} &      \underline{0.076} &      \underline{0.952} &      \textbf{0.061} &      \textbf{0.968} &      \textbf{0.053} &      \textbf{0.979} &      -          & 0.8M \\
\bottomrule
\end{tabular}}
\end{table}

In addition to full-video evaluation, we also report per-frame metrics, where we perform inference on complete videos but calculate metrics (including scale and shift alignment) on individual frames to assess per-frame depth estimation accuracy. The results in Table~\ref{tab:perframesintel} demonstrate that our method outperforms other video depth estimation models in per-frame accuracy, even surpassing Depth Anything V2. This improvement can be attributed to the additional temporal information provided by video context.
To further quantify temporal consistency on long sequences, we evaluate 500-frame ScanNet++~\cite{yeshwanth2023scannet++} videos using Temporal Alignment Error (TAE). As shown in Table~\ref{tab:temporal_consistency}, ICDepth achieves the lowest TAE and the best long-sequence depth accuracy.

\begin{table}[t]
\centering
\caption{Per-frame depth estimation accuracy on Sintel. Metrics are computed on individual frames with per-frame scale-shift alignment.}
\label{tab:perframesintel}
\small
\begin{tabular}{lccc}
\toprule
Method & AbsRel$\downarrow$ & RMSE$\downarrow$ & $\delta_1\uparrow$ \\
\midrule
Depth Anything V2 & 0.226 & 5.121 & 0.740 \\
Video Depth Anything & 0.209 & 4.857 & 0.754 \\
ChronoDepth & 0.431 & 8.565 & 0.603 \\
DepthCrafter & 0.194 & 4.694 & 0.784 \\
Depth Any Video & 0.287 & 5.109 & 0.723 \\
 \textbf{Ours} & \textbf{0.186} & \textbf{4.227} & \textbf{0.787} \\
\bottomrule
\end{tabular}
\end{table}
\begin{table}[t]
\centering
\caption{Temporal consistency and long-sequence evaluation on 500-frame ScanNet++ sequences. TAE denotes Temporal Alignment Error.}
\label{tab:temporal_consistency}
\small
\setlength{\tabcolsep}{3pt}
\begin{tabular}{lcccc}
\toprule
Method & TAE $\downarrow$ & AbsRel $\downarrow$ & RMSE $\downarrow$ & $\delta_1$ $\uparrow$ \\
\midrule
DepthCrafter & 3.07 & 0.204 & 0.658 & 0.589 \\
Depth Any Video & 5.69 & 0.148 & 0.483 & 0.814 \\
Ours & \textbf{2.61} & \textbf{0.131} & \textbf{0.441} & \textbf{0.832} \\
\bottomrule
\end{tabular}
\end{table}

To further assess robustness under unseen distribution shifts, we construct a low-light variant of Sintel by reducing illumination and adding sensor noise. Such low-light videos are not included in our depth training data. As shown in Table~\ref{tab:ood_lowlight}, ICDepth exhibits the smallest relative $\delta_1$ degradation compared with clean Sintel, suggesting better robustness under unseen low-light conditions.

\begin{table}[b]
\centering
\small
\caption{OOD robustness on the unseen low-light Sintel setting. We report the relative $\delta_1$ degradation from clean Sintel to low-light Sintel. Lower is better.}
\label{tab:ood_lowlight}
\begin{tabular}{l c}
\hline
Method & Rel. $\delta_1$ Drop $\downarrow$ \\
\hline
Depth Anything V2 & 13.2\% \\
DepthCrafter & 8.4\% \\
Depth Any Video & 13.2\% \\
\textbf{ICDepth} & \textbf{4.0\%} \\
\hline
\end{tabular}
\end{table}

\subsubsection{Qualitative Comparison}

\begin{figure*}[!ht]
    \centering
    \includegraphics[width=\linewidth]{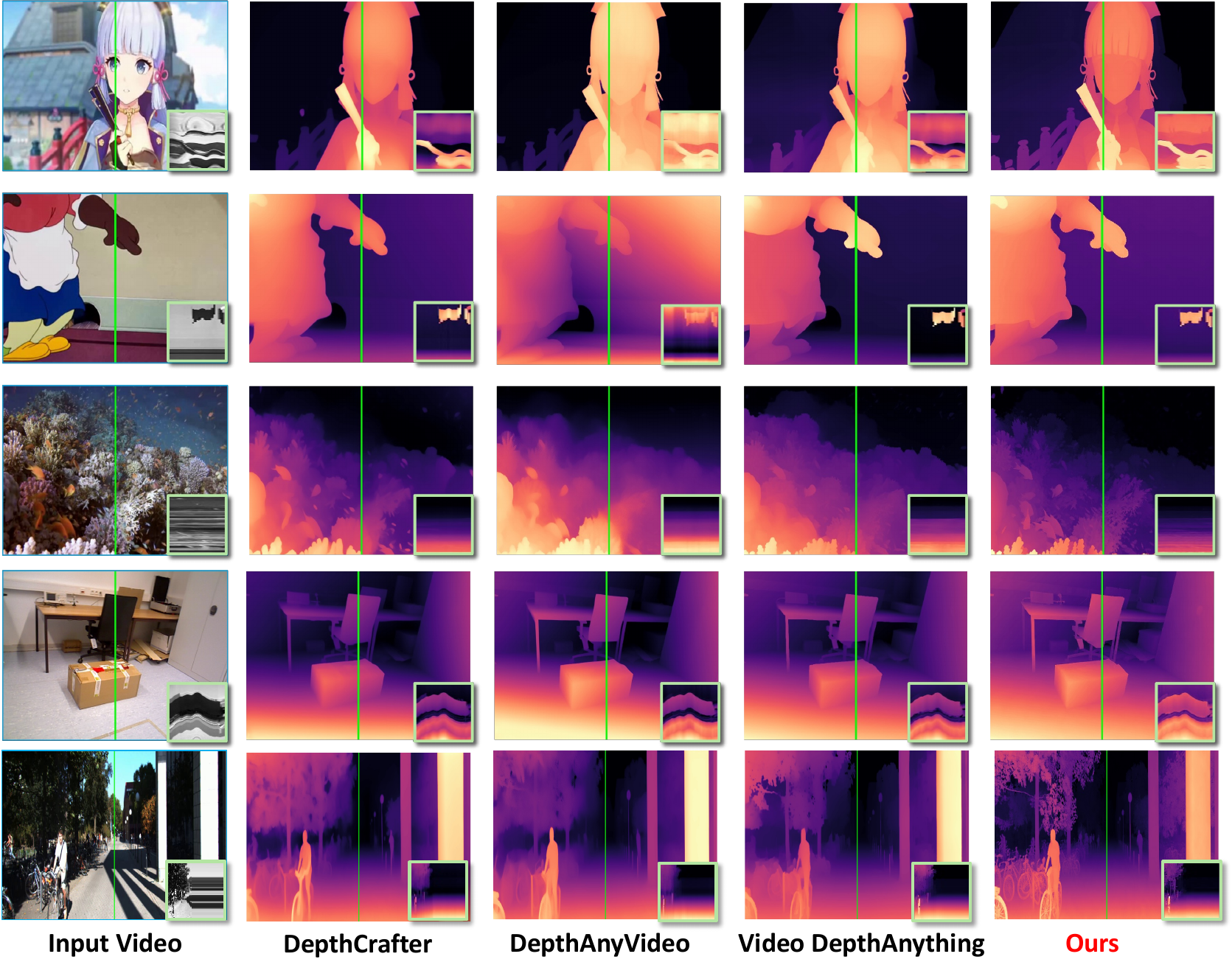}
    \caption{We compare against representative video depth estimation baselines. For temporal consistency visualization, we show depth profiles (green boxes) extracted along the temporal dimension at the green line locations, illustrating the temporal stability of each method.}
    \label{fig:maincomp}
\end{figure*}

\begin{figure}
    \centering
    \includegraphics[width=\linewidth]{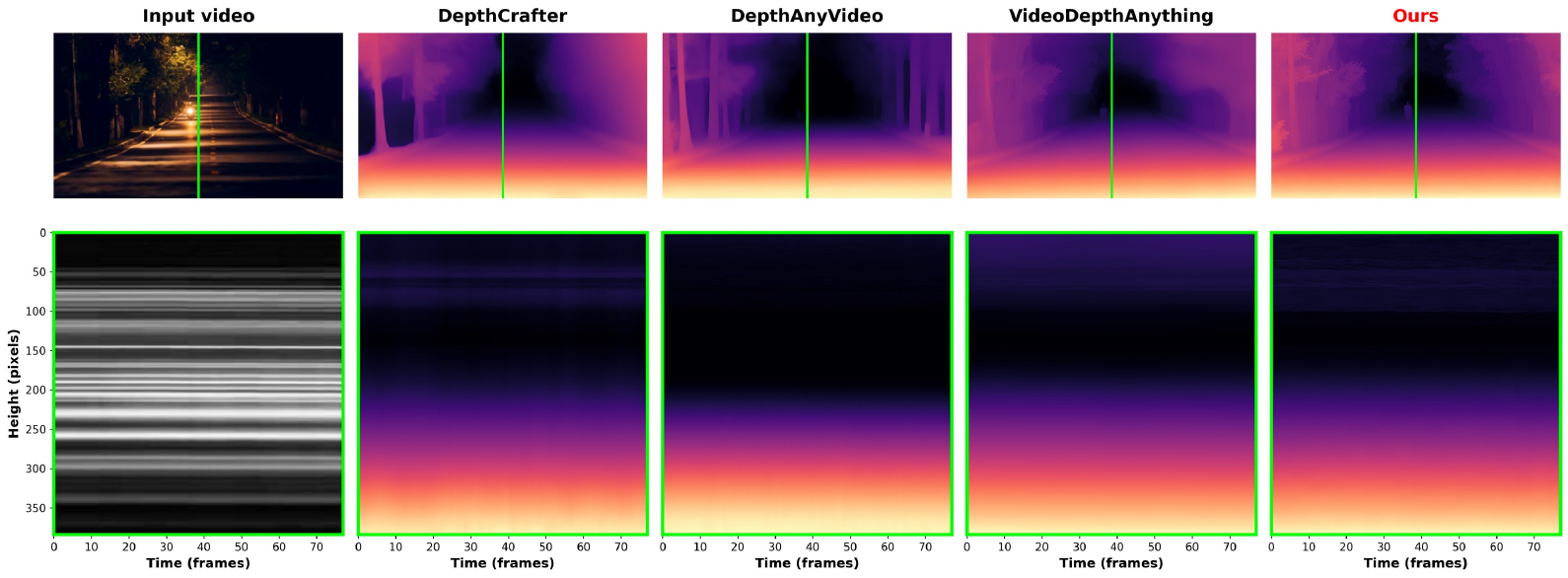}
        \caption{Qualitative comparison under challenging nighttime conditions. We compare our method with state-of-the-art approaches on nighttime scenes. The second row displays spatio-temporal plots along the green scanlines. Our method demonstrates superior depth accuracy and temporal stability, producing flicker-free results even under adverse illumination and visibility.}
    \label{fig:supp_high_res}
\end{figure}

Figure~\ref{fig:maincomp} presents qualitative comparisons across five diverse domains (3D games, 2D cartoons, underwater, indoor, and driving scenes). Our method demonstrates superior capability in producing sharper depth boundaries and recovering fine-grained details consistently missed by competitors. For instance, in underwater scenes (Row 3), we precisely separate fish and coral boundaries, while in driving scenes (Row 5), we uniquely reconstruct bicycles and crowds with clean spatial separation. Temporal analysis (green boxes) further confirms our exceptional consistency without flickering artifacts, validating advantages in boundary precision, temporal coherence, and detail reconstruction. Figure~\ref{fig:supp_high_res} further evaluates depth estimation under challenging low-light conditions. Despite adverse illumination and low visibility in nighttime scenes, our method produces more accurate depth maps.

\subsection{Ablation Studies}
\begin{table}[t]
\centering
\caption{Ablation study of our proposed components conducted on Sintel dataset.}
\label{tab:ablation}
\setlength{\tabcolsep}{1pt} 
\begin{tabular}{lccc}
\toprule
\textbf{Method Variant} & AbsRel $\downarrow$ & RMSE $\downarrow$ & $\delta_1$ $\uparrow$  \\
\midrule
\textbf{Full model (Ours)} & \textbf{0.250} & \textbf{5.155} & \textbf{0.749} \\
\midrule
\multicolumn{4}{l}{\textit{In-Context Conditioning (ICC)}} \\
\quad Replace with channel concat & 0.367 & 5.956 & 0.654 \\
\midrule
\multicolumn{4}{l}{\textit{SAND-Attention}} \\
\quad Replace with full attention & 0.413 & 6.078 & 0.443 \\
\quad w/o RoPE Alignment & 0.410 & 5.999 & 0.450 \\
\quad w/o Decoupled Attention & 0.262 & 5.196 & 0.710 \\
\midrule
\multicolumn{4}{l}{\textit{Semantic-Resolution Aware  feature modulation (SRFM)}} \\
\quad w/o SRFM & 0.306 & 6.325 & 0.696 \\
\quad w/o DINOv2 features & 0.269 & 5.730 & 0.709 \\
\quad w/o Resolution Embedding & 0.264 & 5.565 & 0.728 \\
\bottomrule
\end{tabular}
\end{table}
To validate the effectiveness of our design, we conduct ablation studies on the Sintel dataset to answer three key questions: \textbf{Q1:} \textit{Is ICC superior to standard channel concatenation?} \textbf{Q2:} \textit{How to adapt ICC framework to VDiT-based depth estimation model?} \textbf{Q3:} \textit{Are task-specific priors essential for bridging the generative-perceptive gap?} Results are presented in Table~\ref{tab:ablation}.

\noindent\textbf{Q1: ICC vs. Channel Concatenation.}
We compare our ICC framework against the standard adaptation approach using Wan2.1 Control~\cite{alibaba-pai-wan2.1-fun-1.3b-control}, where RGB and depth features are concatenated along the channel dimension and processed through a modified input projection layer.
Table~\ref{tab:ablation} shows ICC dramatically outperforms channel concatenation. This validates that ICC's token-level conditioning is crucial: the VDiT's 
attention mechanism can directly model RGB-depth correspondences at matching spatial-temporal locations, rather than relying on implicit late-stage fusion through separate channels.

\noindent\textbf{Q2: Critical Modifications to ICC framework.}
 We conduct ablation studies to identify essential modifications for adapting the ICC framework to VDiT-based depth estimation, evaluating three variants of our proposed SAND-Attention mechanism.

\textit{Naive full attention.} Replacing SAND-Attention with standard full attention (lacking both RoPE Alignment and Decoupled Attention) results in {complete performance degradation}, demonstrating the fundamental inadequacy of the original ICC framework for this task.

\textit{RoPE Alignment.} Removing only the shared RoPE alignment also causes severe failure, confirming that establishing precise spatial-temporal correspondence between RGB and depth tokens at shared $(x,y,t)$ coordinates is paramount. Without this explicit alignment, the model cannot learn proper cross-modal relationships.

\textit{Decoupled Attention.} Removing the attention decoupling mechanism yields moderate but significant degradation (AbsRel: 0.262 vs. 0.250). The substantial drop in $\delta_1$ (0.710 vs. 0.749) particularly highlights how bidirectional attention flow corrupts confident predictions through noise contamination, though this effect proves less critical than spatial misalignment.

\noindent\textbf{Q3: Necessity of Task-Specific Priors.}
We conduct ablation studies to investigate the necessity of semantic and resolution information injection, evaluating three variants of our SRFM module.

\textit{Full SRFM removal.} Complete removal of SRFM causes substantial performance degradation, demonstrating that T2V generative features alone are insufficient for precise depth estimation. The magnitude of this degradation exceeds that of any individual SRFM component, revealing synergistic effects between DINOv2 semantic priors and resolution embeddings.

\textit{DINOv2 features.} Removing DINOv2 features results in moderate performance decline. This validates that DINOv2's rich semantic understanding significantly enhances geometric reasoning capabilities.

\textit{Resolution Embeddings.} The ablation of resolution embeddings yields a measurable performance drop, confirming their role in providing complementary spatial context. This information proves particularly critical on datasets with unconventional resolutions, such as KITTI, where explicit resolution conditioning enhances the model's robustness and generalization capability to diverse and unseen spatial dimensions.

\subsection{Efficiency Analysis}
We compare the computational overhead of our method against both discriminative and generative video depth estimation models in Table~\ref{tab:efficiency}. While discriminative feed-forward models such as Video Depth Anything are naturally faster, ICDepth remains efficient within the generative diffusion paradigm. Compared with Depth Any Video, ICDepth achieves comparable inference speed while substantially reducing GPU memory consumption, obtaining stronger depth estimation accuracy. 

\begin{table}[h]
\centering
\small
\setlength{\tabcolsep}{3pt}
\caption{Efficiency at resolution $480\!\times\!640$, 53 video frames.}
\label{tab:efficiency}
\begin{tabular}{l|c|ccc}
\toprule
Method & Type & Time (s)$\downarrow$ & FPS$\uparrow$ & Mem (GB)$\downarrow$ \\
\midrule
Video Depth Anything & Discr. & 1.82 & 29.1 & 8.5 \\
Depth Any Video  & Gen. & 11.19 & 4.7 & 33.0 \\
\textbf{Ours} & Gen. & 11.85 & 4.5 & \textbf{11.0} \\
\bottomrule
\end{tabular}
\vspace{-0.6em}
\end{table}
Moreover, as shown in Table~\ref{tab:sampling_steps}, ICDepth achieves a strong accuracy--efficiency balance with only 3--5 sampling steps.
\begin{table}[h]
\centering
\small
\caption{Quantitative results with different sampling steps on Sintel.}
\label{tab:sampling_steps}
\begin{tabular}{cccc}
\toprule
\textbf{Steps} & \textbf{AbsRel}$\downarrow$ & \textbf{RMSE}$\downarrow$ & \textbf{$\delta_1$}$\uparrow$ \\
\midrule
3  & 0.262 & 5.330 & 0.746 \\
5  & 0.250 & 5.154 & \textbf{0.749} \\
10 & 0.229 & 4.584 & 0.735 \\
20 & \textbf{0.217} & \textbf{4.469} & 0.721 \\
\bottomrule
\end{tabular}
\end{table}

\section{Conclusion}
We introduce ICDepth, a novel framework adapting pre-trained T2V diffusion transformers via In-Context Conditioning for video depth estimation. Through SAND-Attention and SRFM (tackling geometric ambiguity), our method preserves pre-trained spatial-temporal priors while adding perceptual precision.

ICDepth achieves state-of-the-art results on multiple benchmarks with remarkable data efficiency: trained on only 0.8M frames (6-13$\times$ less than competing generative methods), it delivers superior temporal consistency, per-frame accuracy, and strong zero-shot generalization to diverse domains. This demonstrates that foundation model priors can be effectively transferred to dense prediction tasks through careful architectural design.



\section*{Acknowledgments}
The work was supported by the Research Grants Council of HKSAR under grant number AoE/E-601/24-N.

%
%
\bibliographystyle{splncs04}
\bibliography{main}
\end{document}